\documentclass{article}

\PassOptionsToPackage{numbers, compress}{natbib}
\usepackage[preprint]{neurips_2020}

\usepackage[utf8]{inputenc} 
\usepackage[T1]{fontenc}    
\usepackage{hyperref}       
\usepackage{url}            
\usepackage{booktabs}       
\usepackage{amsfonts}       
\usepackage{nicefrac}       
\usepackage{microtype}      
\usepackage{graphicx}
\usepackage{paralist}
\usepackage{xcolor}
\usepackage{enumitem}
\usepackage{verbatim}

\title{On the Reliability and Generalizability of Brain-inspired Reinforcement Learning Algorithms}

\author{%
  Dongjae Kim\(^*\) \\
  Department of Bio and Brain Engineering\\
  KAIST\\
  Republic of Korea \\
  \texttt{kim10481@kaist.ac.kr} \\
  \And
  Jee Hang Lee\thanks{These authors contributed equally. \(\dagger\) Corresponding author.} \\
  Department of Human-Centered AI \\
  SangMyung University, Seoul \\
  Republic of Korea \\
  \texttt{jeehang@smu.ac.kr}
  \And
  Jae Hoon Shin,~Minsu Abel Yang,~Sang Wan Lee\(^{\dagger}\) \\
  Department of Bio and Brain Engineering\\
  KAIST\\
  Republic of Korea \\
  \texttt{\{skalclrptsp, minsuyang, sangwan\}@kaist.ac.kr} \\
}

\begin{document}

\maketitle

\begin{abstract}
Although deep RL models have shown a great potential for solving various types of tasks with minimal supervision, several key challenges remain in terms of learning rapidly from limited experience, adapting to environmental changes, and generalizing learning from a single task. Recent evidence in decision neuroscience has shown that the human brain has an innate capacity to resolve these issues, leading to optimism regarding the development of neuroscience-inspired solutions toward sample-efficient, adaptive, and generalizable RL algorithms. We show that the computational model, adaptively combining model-based and model-free control, which we term the prefrontal RL, reliably encodes the information of high-level policy that humans learned, and this model can generalize the learned policy to a wide range of tasks. First, we trained the prefrontal RL, deep RL, and meta RL algorithms on 82 human subjects’ data, collected while human participants were performing two-stage Markov decision tasks, in which we experimentally manipulated the goal, state-transition uncertainty, and state-space complexity. In the reliability test, which is based on a combination of the latent behavior profile and the parameter recoverability test, we showed that the prefrontal RL reliably learned the latent policies of the human subjects, while all the other models failed to pass this test. Second, to empirically test the ability to generalize what these models learned from the original task, we situated them in the context of environmental volatility. Specifically, we ran large-scale simulations with 10 different Markov decision tasks, in which latent context variables change over time. Our information-theoretic analysis showed that the prefrontal RL showed the highest level of adaptability and episodic encoding efficacy. To the best of our knowledge, this is the first attempt to formally test the possibility that computational models mimicking the way the brain solves general problems can lead to practical solutions to key challenges in machine learning.

\end{abstract}

\section{Introduction}
\label{sec:introduction}

\textbf{Fundamental challenges for reinforcement learning (RL)}. Rapid advances in reinforcement learning (RL) have offered great potential for developing algorithms to solve various types of complex problems \cite{mnih2015human,silver2016, silver2017a,silver2018general,vinyals2019grandmaster}. For example, hierarchical architectures have been shown to promote efficient exploration with sparse rewards~\cite{kulkarni2016hierarchical,hamrick2017metacontrol}. Model-based RL has demonstrated its ability to improve sample efficiency in many situations~\cite{guo2014deep, silver2016, racaniere2017imagination, foerster2018learning, janner2019trust}. RL algorithms have also established biological relevance \cite{wang2016learning, wang2018meta, dabney2020distributional,barreto2017successor}, increasing optimism about the building of models with human-like intelligence. Despite their capacity to solve a variety of tasks, several key challenges remain, such as improving sample efficiency, adaptability, and generalization. For example, RL algorithms lack the ability to rapidly learn the structure of the environment. Moreover, their behavioral policy is often highly biased, making it hard to adapt to changing environments or transfer their task knowledge to general situations~\cite{lake2017building}.
 
\textbf{Brain’s solutions to RL.} Earlier studies showed that value-based decision-making is guided by reward prediction error (RPE), and the midbrain dopamine neurons encode this information~\cite{montague1996, schultz1997}. A later study found that the human brain appears to implement an actor-critic scheme~\cite{o2004dissociable}. These studies support the idea that the way the brain learns from experience bears a resemblance to model-free RL. That being said, a single model-free RL can account for relatively small variability in behavior and neural data. This conventional view was then challenged by the idea that the brain implements more than one type of RL~\cite{Daw2005}. Indeed, the human brain is capable of not only combining model-free and model-based RL~\cite{daw2011model} but also adaptively choosing one strategy over the other depending on the context changes ~\cite{glascher2010states, lee2014neural}. This adaptive process was found to be guided by a part of the lateral prefrontal cortex, which compiles the reliability of respective predictions made by the model-free and model-based RL strategy~\cite{lee2014neural, kim2019task}. The brain also has a propensity for pursuing a computationally less expensive strategy, such as model-free RL, especially in a highly stable or volatile environment. On the other hand, the prefrontal cortex engages in drastically improving the sample efficiency of model-based learning by compromising performance reliability~\cite{lee2015neural}. This implies that the brain has an innate ability to deal with the tradeoff between performance, sample efficiency, and computational cost~\cite{lee2019}. Critically, it leads to the theoretical implication that the brain explores learning strategies in a way that best responds to new challenges in the environment.
 
\textbf{Can RL algorithms learn from the brain’s RL?} There are a few commonalities between the brain and algorithmic solutions to adaptive RL, but the substantial difference still lies in the way they approach problems. Moreover, the capacity of the brain to effectively deal with the challenges of RL has not been fully developed by RL algorithms. This raises the following interesting questions: Is it possible for the RL models to glean information about human RL directly from human behavioral data? Then, do these imitation models have a similar policy as humans? While many works have successfully demonstrated the effectiveness of policy learning from imitation~\cite{duan2017one, ding2019goal, henderson2018optiongan}, little is known about whether their policies are similar to humans’ latent policy or whether a policy can be generalized to other tasks. 
Another potential issue is overfitting. Notably, recent studies examining the recoverability of human behavior~\cite{broomell2014parameter, evans2020parameter} have shown that models often fail to replicate the findings based on human behavior data, to which they are originally fitted. This suggests that the learned behavioral policy of computational models does not fully reflect the innate dynamics of human RL.  
 
\textbf{Main contributions of our work.} This study examines the following fundamental question: Is it possible for algorithms to learn generalizable policies from humans? To this end, we devide this problem into two formal tests as prerequisites: reliability test and empirical generalizability test. Our works are summarized as follows.

\begin{itemize}[leftmargin=*]
\item Learning human latent policy (Section 2). We fitted 82 human subjects’ data (Figure 1-(A)) to different types of RL models, each of which implements model-free and model-based control in different ways, including deep RL, meta RL, and prefrontal RL (Figure 2). We used data collected from human participants performing two-stage Markov decision tasks, in which the goal, state-transition uncertainty, and state-space complexity were experimentally manipulated.
\vspace{-0.3em}

\item Reliability test (Section 3). Using rigorous latent behavior profile recoverability tests (Figure 1–(B)), we show that the latent policy of the computational model, adaptively combining model-based and model-free control, called prefrontal RL, is qualitatively similar to that of human subjects, whereas all other models fail to replicate the effect (Figure 3)..
\vspace{-0.3em}

\item Empirically generalizability test (Section 4). To test the model’s ability to generalize what it learned from the original task (Figure 1-(C)), we ran large-scale simulations with 10 different Markov decision tasks in which the latent context variables change over time (Figure 4). We found that the prefrontal RL showed the highest level of adaptability (Figure 5) and episodic encoding efficacy (Figure 6).
\vspace{-0.3em}

\item Implications. Our results carry broad implications. This work is the first attempt to formally test the possibility that computational models can reliably learn the latent policy of humans. Moreover, our approach can offer practical solutions to key challenges in machine learning, enabling the design of more human-like intelligence. 
\end{itemize}


\section{Learning latent policy of humans}
\label{sec:experimental framework}

\begin{figure}[t!]
  \centering
  \includegraphics[width=\textwidth]{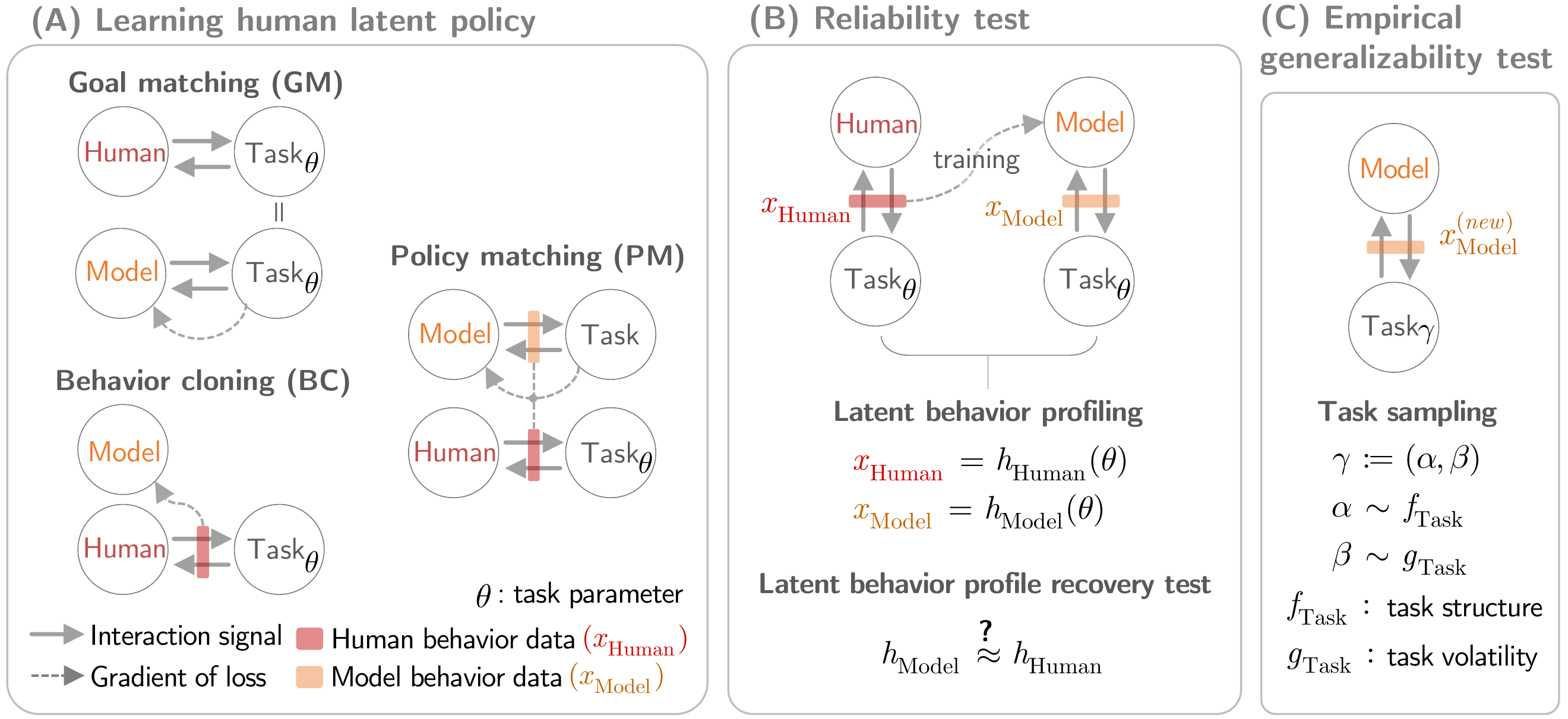}
  \caption{General framework. (A) Outline of three different approaches to training the RL models: goal matching (GM), behavior cloning (BC), and policy matching (PM). PM is a blend of GM and BC, used in~\cite{lee2014neural, kim2019task}. This is intended to mimic the way humans perform reward maximization. (B) \textit{Reliability test} outlines our validation process, which compares the latent behavioral profile of human subjects and that of the RL models. After completing the initial \textit{training} with the original human data (\(x_{Human}\)), we obtain the simulation data (\(x_{Model}\)) by having the fitted RL models (\texttt{Model}) perform the original task. We then quantify the effect of the task parameters on the human behavioral data and the simulation data, respectively (``latent behavior profiling''), and test whether the effects of the two cases match.. (C) \textit{Empirical generalizability test.} We generate new tasks by sampling task parameters from some task distributions ("task sampling"), and examine the performance of each RL models .}
  \label{fig:framework}
  \vspace{-0.5em}
\end{figure}

In order to build RL models that learn and perform tasks in a similar way as humans, we consider three training methods: goal matching (\textit{GM}), behavior cloning (\textit{BC}), and policy matching (\textit{PM}). This process, which we call human latent policy learning, is intended to learn behavioral policy directly from human behavioral data.

\textbf{Goal matching.} The RL model interacts with the task environment to maximize the expected amount of future reward, so it does not use any human behavior data for training. However, the task (goal) used for training the model is exactly the same as the one performed by human subjects (Figure 1A-GM). For this reason, we call this method goal matching.

\textbf{Policy matching.} Policy matching combines the goal matching and behavioral cloning, making it possible to achieve both goal matching and behavior matching. Specifically, the RL model is trained in such a way that it mimics the way the human performs reward maximization. In each training epoch, the RL model completes one episode of the task to maximize reward (goal matching), and then the difference between the behavior of the model and that of the human subjects is translated into the loss function (behavior cloning). This method was previously used for training computational models to account for neural data~\cite{glascher2010states, lee2014neural, kim2019task}. 
Note that we do not consider a standard inverse RL method, because it is not directly applicable to the tasks with rapid context changes~\cite{lee2014neural}. Indeed, it is almost impossible for the inverse RL method to estimate our reward function, in which both the reward value and the environmental statistics change over time, and the sample size is too small (around 400 trials per subject).

\begin{figure}[t!]
  \centering
  \includegraphics[width=\textwidth]{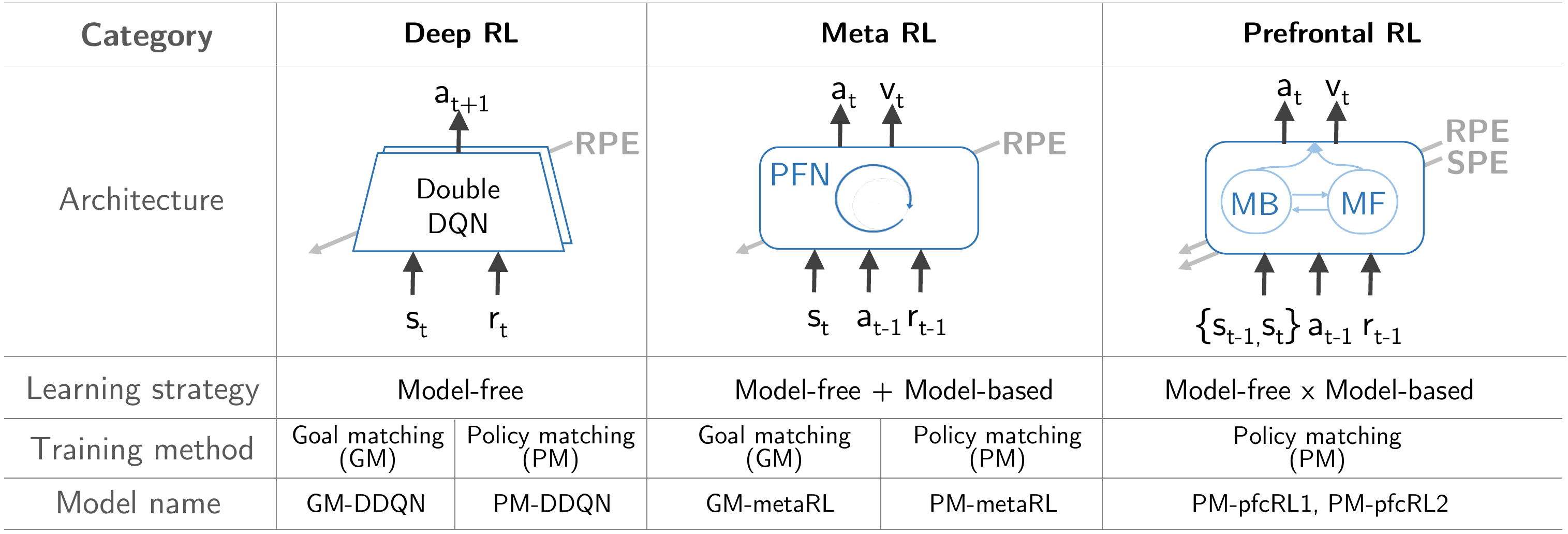}
  \caption{RL models used in our experiments. Three types of RL were implemented using six different RL models: deep RL (GM-DDQN and PM-DDQN), meta RL (GM-metaRL and PM-metaRL), and prefrontal RL (PM-pfcRL1 and PM-pfcRL2). For more details, refer to \textit{Supplementary Methods.}}
  \label{fig:rl models}
  \vspace{-0.5em}
\end{figure}

\textbf{RL models.} We used three different types of RL models: deep RL~\cite{van2016}, meta RL~\cite{wang2016learning, wang2018meta} and prefrontal RL~\cite{lee2014neural, kim2019task}. Figure 2 shows the architecture of each type of model. The first type was implemented with Double DQN (\textit{deep RL})~\cite{van2016}, also known as DDQN. It is one of the typical deep RL models approximating model-free RL. We used both the goal matching and policy matching methods to train this model (GM-DDQN and PM-DDQN, respectively). The second type was implemented with meta RL (\textit{meta RL})~\cite{wang2016learning, wang2018meta} . This model accommodates both model-free and model-based RL. In particular, meta RL is known to adaptively respond to contextual changes in the environments. We used both the goal matching and policy matching methods to train this model (GM-metaRL and PM-metaRL, respectively).

The third type of RL model was implemented with the computational model to account for the neural activity of the lateral prefrontal cortex and ventral striatum (\textit{prefrontal RL})~\cite{lee2014neural, kim2019task}. There are two versions of this model: the baseline model~\cite{lee2014neural} and the adaptive model~\cite{kim2019rldm}. These models learn a task by dynamically arbitrating between model-free and model-based RL. Specifically, they adjust on a trial-by-trial basis the degree of control allocated to the model-free and model-based RL strategies, and this top-down control signal is computed based on the prediction reliability of each RL strategy. We used the policy matching method to train these two models (PM-pfcRL1 and PM-pfcRL2, respectively). We did not use goal matching in this case because previous studies have found that this method is not effective in fitting these models to data ~\cite{lee2014neural}.

\section{Reliability of brain-inspired RL models}
\label{sec:reliability}

\subsection{Recoverability of the latent behavior profile}
\label{subsec:recoverability}

To assess to what extent the RL models reliably learn to mimic human behavior and latent policy, we conducted a \textit{reliability test} (Figure 1-(B)). The test validates the capacity to encode the information of high-level policy that the humans learned while performing the task. The process consists of latent behavior profiling and a recovery test.

\textbf{Latent behavior profiling.} One general way to assess the latent policy that humans learn from a task is to quantify the effect of the latent task parameters (e.g., goal and state-transition uncertainty) on behavior. This measure reflects how the learning agent changes its behavior in response to the change in the environment structure. For the given task parameter \(\theta\) and behavioral data \(x\),  respectively, the latent behavior profile \(h\) is defined as follows:
\[x=h(\theta_{Task}),\]
where \(h\) can be any parameterized function, such as a polynomial function or a neural network. If the task performance of the agent is independent of context changes or if the agent makes random choices, then the effect size (i.e., parameter values of \(h\)) would be zero. In this study, we simply use a general linear model as \(h\).

\textbf{Latent behavior profile recovery test.} The purpose of this test is to evaluate the consistency between the latent policy of the human and that of the RL model. After fitting the model's parameters to the human subjects' data \(x_{Human}\), we generate the simulated data \(x_{Model}\)  by running simulations with the original fitting model on the original task. We then conduct the latent behavior profiling on \(x_{Human}\) and \(x_{Model}\) , respectively. A significant positive correlation between these two latent profiles indicates that the latent policy that the RL model learned is similar to the latent policy of the human.

\begin{figure}[t!]
  \centering
  \includegraphics[width= \textwidth]{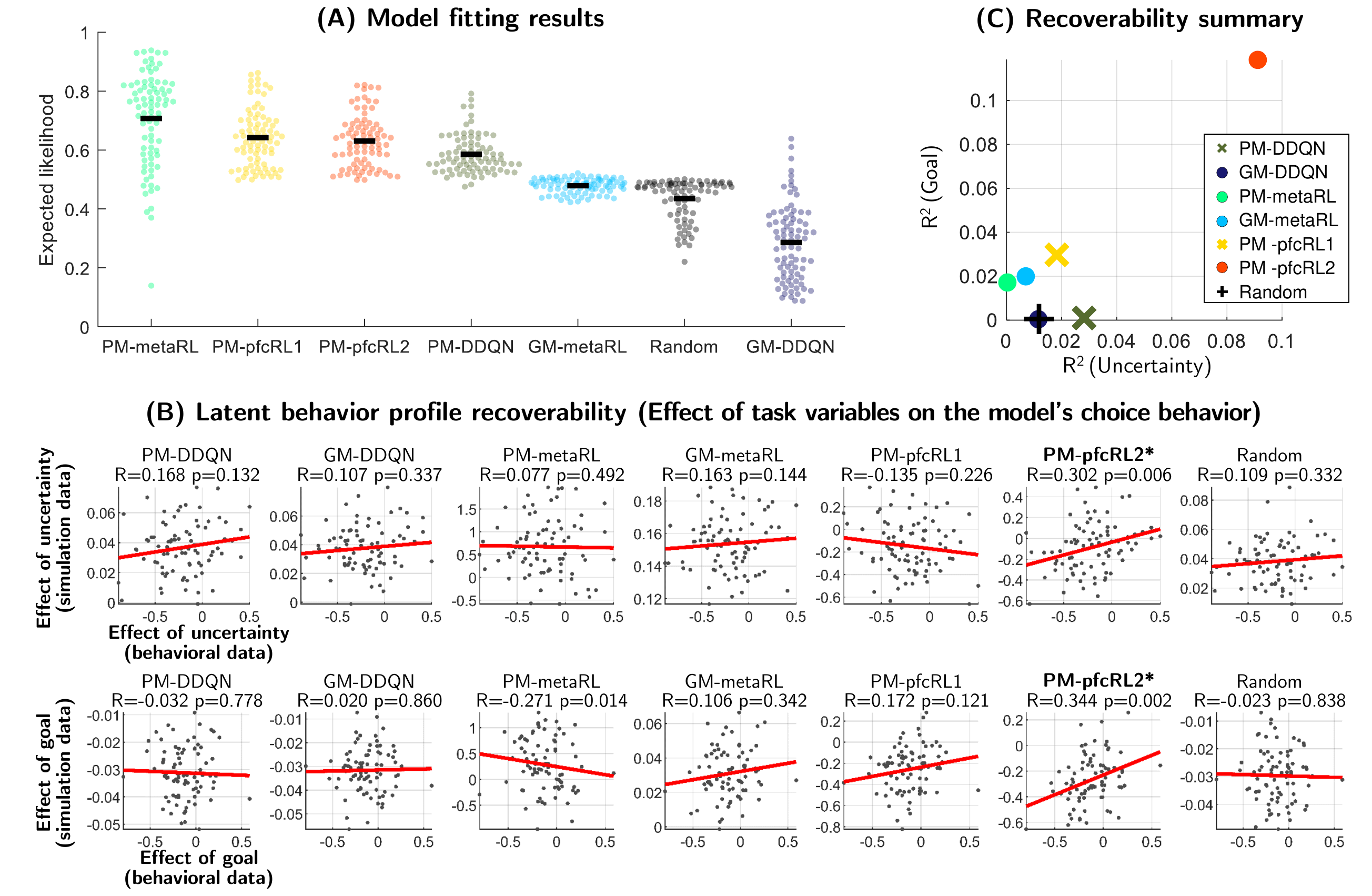}
  \caption{Reliability test results. (A) Model fitting results. (B) Latent behavior profile recoverability. For this, we compared the latent behavior profile computed from the human subjects' data and that from the simulation data (Figure 1-(B)). We evaluated the effect of two task variables, \textit{goal} and \textit{uncertainty}, on the agent's choice behavior. The X-axis and Y-axis show the effect of the task variable measured in the behavioral data and simulation data, respectively. (C) Summary statistics of (B). The X-axis and Y-axis refers to the \(R^2\) of the effect on the agent's choice behavior regarding uncertainty and the goal, respectively  (also see \textit{Supplementary Methods}).}
  \vspace{-0.5em}
\end{figure}

\subsection{Reliability test}
\label{subsec:reliability test}

\textbf{Experimental procedure.} In order to examine the recoverability of the latent behavior profile, we conducted a series of experiments with six different RL models (described in Figure 2) and a random agent as a control condition. In the first step, we trained \textit{prefrontal RL}, \textit{meta RL}, and \textit{deep RL} on 82 human subjects' data (\(x_{Human}\) in Figure 1-(B)). The dataset was collected while the human participants performed two-stage Markov decision tasks. In the second step, we collected another behavioral dataset (\(x_{Model}\) in Figure 1-(B)) by running another set of simulations in which all the RL models performed the same two-stage Markov decision task. We then computed the latent behavior profile \(h_{Human}\), \(h_{Model}\)  as follows:
\[x_{Human}=h_{Human}(\theta_{Task}), x_{Model}=h_{Model}(\theta_{Task}),\]
where \(\theta_{Task}\) represents the task parameters. This is a large-scale experiment, including more than 1,000 model fitting processes: 7 (models) \(\times\) 82 (subjects) \(\times\) 2 (training and retraining).

\textbf{Latent behavior profile recoverability of RL models.} Figure 3-(A) shows the simulation results. In terms of model fitting that quantifies behavior matching between RL models and human subjects, PM-meta RL showed the highest performance, followed by \textit{prefrontal RL} and \textit{deep RL}. As expected, the RL models trained with goal matching showed relatively poor fitting performance. 

However, in the systematic recovery analysis of the latent behavior profiles, we found that the latent behavior profile of the \textit{prefrontal RL} model (PM-pfcRL2) was qualitatively similar to that of the human subjects, whereas all the other RL models failed to replicate the effect (Figure 3-(B)). Although \textit{meta RL} trained with the \textit{PM} method showed significant correlation in some cases, the correlation is negative, indicating that the way this model performs the task may be fundamentally different from that of humans. When computing goodness-of-fit statistics that take into account both the steepness and the significance of the correlation, this effect becomes more dramatic (Figure 3-(C)). The effect size of the \textit{Prefrontal RL} model (PM-pfcRL2) is more than three times larger than the effect sizes of all the other RL models. These results suggest that simply imitating human behavior (as shown in Figure 3-(A)) does not necessarily mean that the agent actually learns the latent policy of the human (as shown in Figure 3-(B) and 3-(C)).

\section{Empirical genralizability of brain-inspired RL models}
\label{sec:generalizability}

\subsection{Markov decision tasks with varying degree of volatility}
\label{subsec: mdp with volatility}

\begin{figure}[t!]
  \centering
  \includegraphics[width=\textwidth]{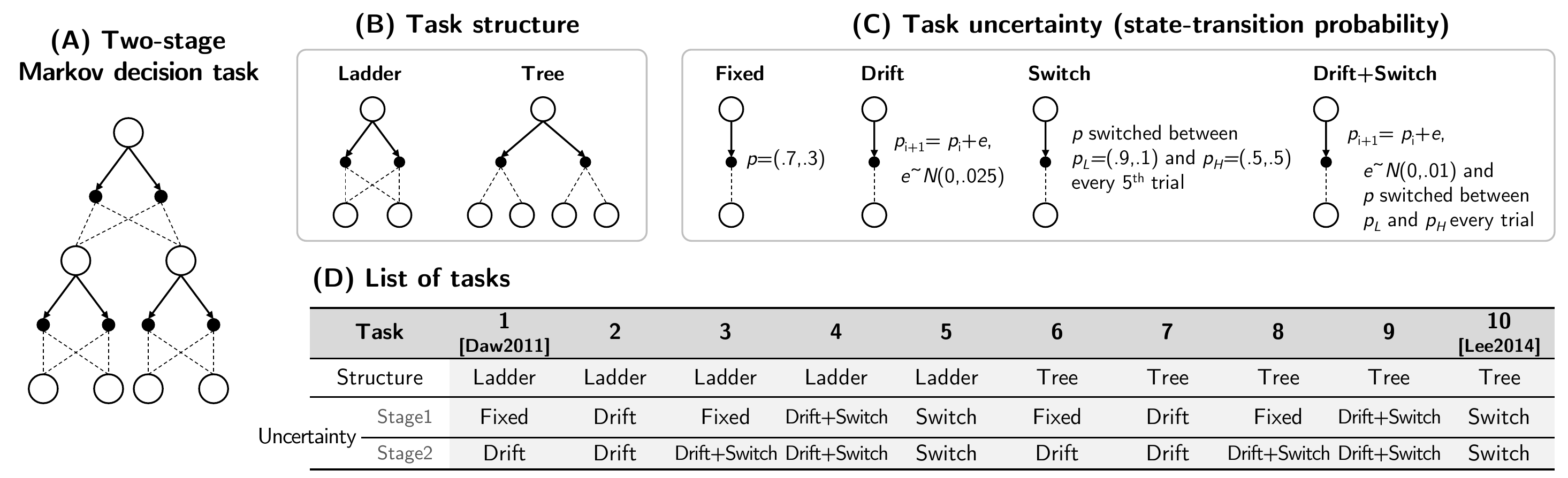}
  \caption{Simulation environments for testing the generalizability of each RL model. (A) An example showing the two-stage Markov decision task. (B) Task structure conditions. (C) Task uncertainty conditions. (D) A list of tasks. We created a total of 10 different types of tasks by systematically manipulating both task structure and task uncertainty (also see \textit{Supplementary Methods}).}
  \label{fig:env}
  \vspace{-0.5em}
\end{figure}

To empirically test the models' capacity to generalize from what they learned from the original task to other tasks (Figure 1-(C)), we situated them in the context of environmental volatility. Using the same set of RL models as in Sections 2 and 3, we ran large-scale simulations with 10 different Markov decision tasks, each of which manipulated latent context variables in different ways (Figure 4). 

The tasks were created by systematically manipulating two task parameters: task structure (ladder and tree) and task uncertainty (fixed, drift, switch, and drift + switch). For the task structure, we used a ladder and tree type (Figure 4-(B)). For the task uncertainty change, we considered four different types of state transition functions (Figure 4-(C)), each of which changed, on a trial-by-trial basis, the state-transition probability values in a different manner. The first type (``fixed'') uses a fixed state-transition probability. The second type (``drift'') uses the state-transition probability following random walks, in which the state-transition probability value changes relatively slowly. The third type (``switch'') alternates between two different state-transition conditions: conditions with low and high uncertainty, respectively. In this task, the learning agent experiences abrupt changes in the task structure and needs to adapt quickly. The fourth type (``drift + switch'') is a mixture of the second and third types. The full configurations of each task are provided in Figure 4-(D). Note that Task 1 and Task 10 correspond to tasks used in previous studies investigating the brain's RL processes~\cite{daw2011model, lee2014neural}. 

\subsection{Adaptability and generalizability}
\label{subsec: adaptability}

\begin{figure}[t!]
  \centering
  \includegraphics[width=\textwidth]{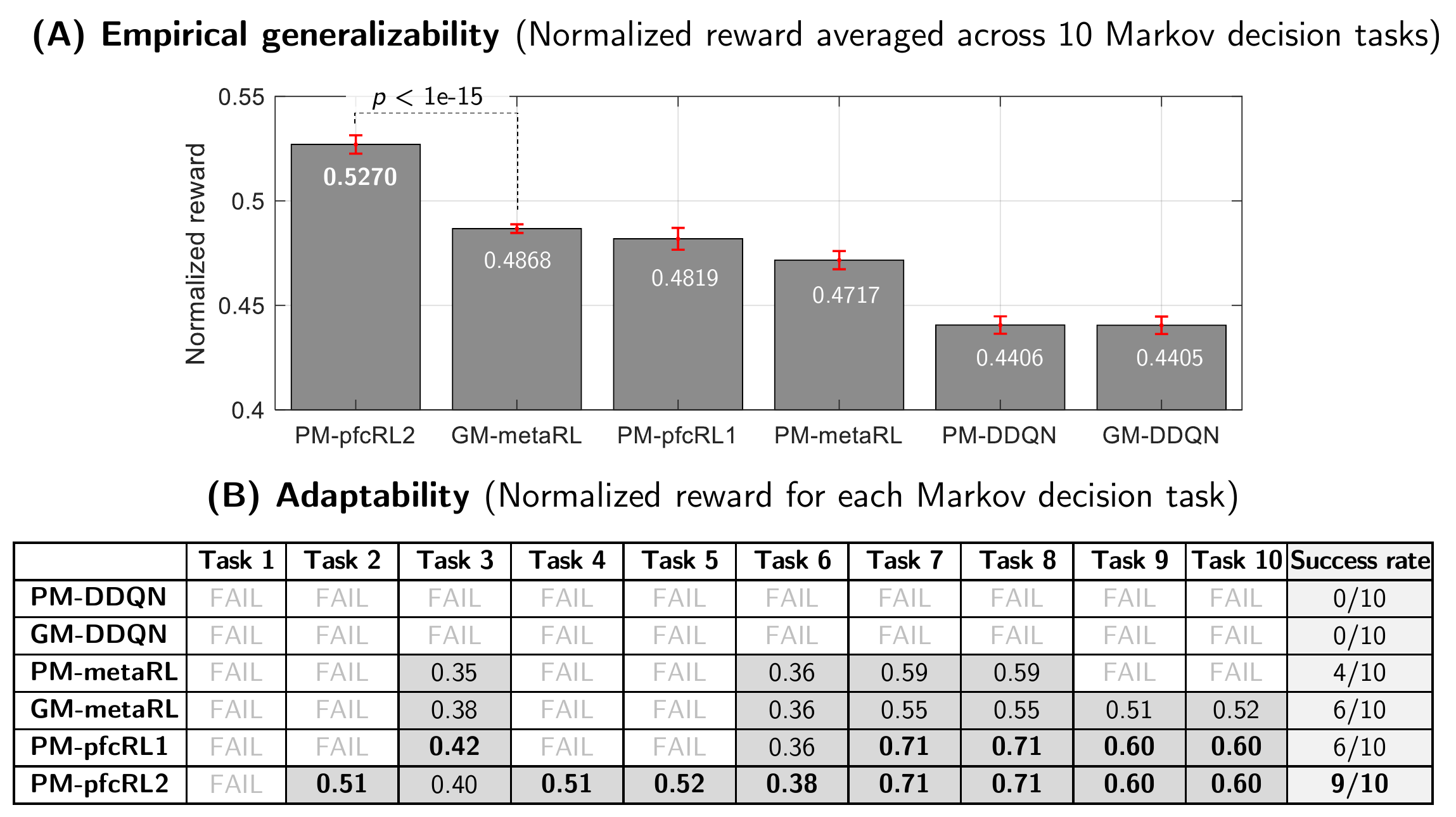}
  \caption{Simulation results on the adaptability of the RL models. (A) Normalized reward averaged across 10 tasks. It quantifies the models' capacity to generalize what they learned in the two-stage Markov decision task~\cite{lee2014neural} to other tasks. (B) Normalized reward in each task. ``FAIL'' means that the corresponding RL model failed to outperform the random agent (paired \(t\)-test; \(p\) > 0.05).}
  \label{fig:adaptability}
  \vspace{-0.5em}
\end{figure}
To test empirical generalizability, we ran simulations in which the six RL models, trained on the original dataset (used in Sections 2 and 3), performed 10 Markov decision tasks (described in Section 4.1). This involved 4,920 simulations (= 82 subjects \(\times\) 6 RL models \(\times\) 10 tasks) in total. Note that the average performance across all the tasks represents the empirical generalizability, and the performance on each task represents the adaptation ability of the corresponding model in different situations.

We found that the \textit{Prefrontal RL} model showed the highest level of generalizability (Figure 5-(A)). Notably, the PM-pfcRL2 successfully solved nine tasks out of ten and scored the highest on eight tasks out of nine in terms of the normalized reward (Figure 5-(B)). Both the GM-metaRL and PM-pfcRL1 showed the second-best performance. Although the performance of the PM-pfcRL1 was the same as that of the GM-metaRL, the PM-pfcRL1 outperformed in five out of six tasks. Taken together, these results suggest that the Prefrontal RL models (PM-pfcRL1 and PM-pfcRL2) have the best ability to generalize what they learn from the original task. 

\subsection{Episodic encoding efficacy}
\label{subsec:episodic encoding}

\textbf{Potential information-theoretic measure for quantifying generalizability of RL models.} To better understand the nature of the ability to generalize, we conducted an information-theoretic analysis. This analysis is designed to quantify (1) the amount of information transferred from the observation of the past episodes of events to the RL model's action and (2) the degree of optimality in its action. We hypothesized that the higher the generalizability, the more efficiently the RL model encodes the episodic information to generate optimal action. As such, we expect that the generalizability of the model can be quantified as (1) the mutual information from the episodic events and the agent’s action~\cite{filipowicz2020complexity} (``episodic encoding efficiency'') as well as (2) the mutual information of the agent's action and the optimal action (``choice optimality''). The optimal action was defined as the action taken by the ideal agent, assuming that it is fully informed on the task's parameter changes. The episodic encoding efficiency is defined as \(I(F_{t-1};a_t)\), where \(F_{t-1}\) and \(a_t\) are the episode variable at trial \(t-1\) and the action at trial \(t\), respectively. The choice optimality is defined as \(I(a_t;a_t^*)\), where \(a_t\) and \(a_t^*\) are the choices (actions) of the RL agent and ideal agent, respectively.

We hypothesized that one fundamental requirement of a highly generalizable RL agent is the ability to transfer information from past episodes to its action and task performance. Accordingly, the correlation between episodic encoding efficiency and choice optimality, called ``episodic encoding efficacy'', can be one potential information-theoretic indicator of the generalizability of the RL model.

\begin{figure}[t!]
  \centering
  \includegraphics[width=\textwidth]{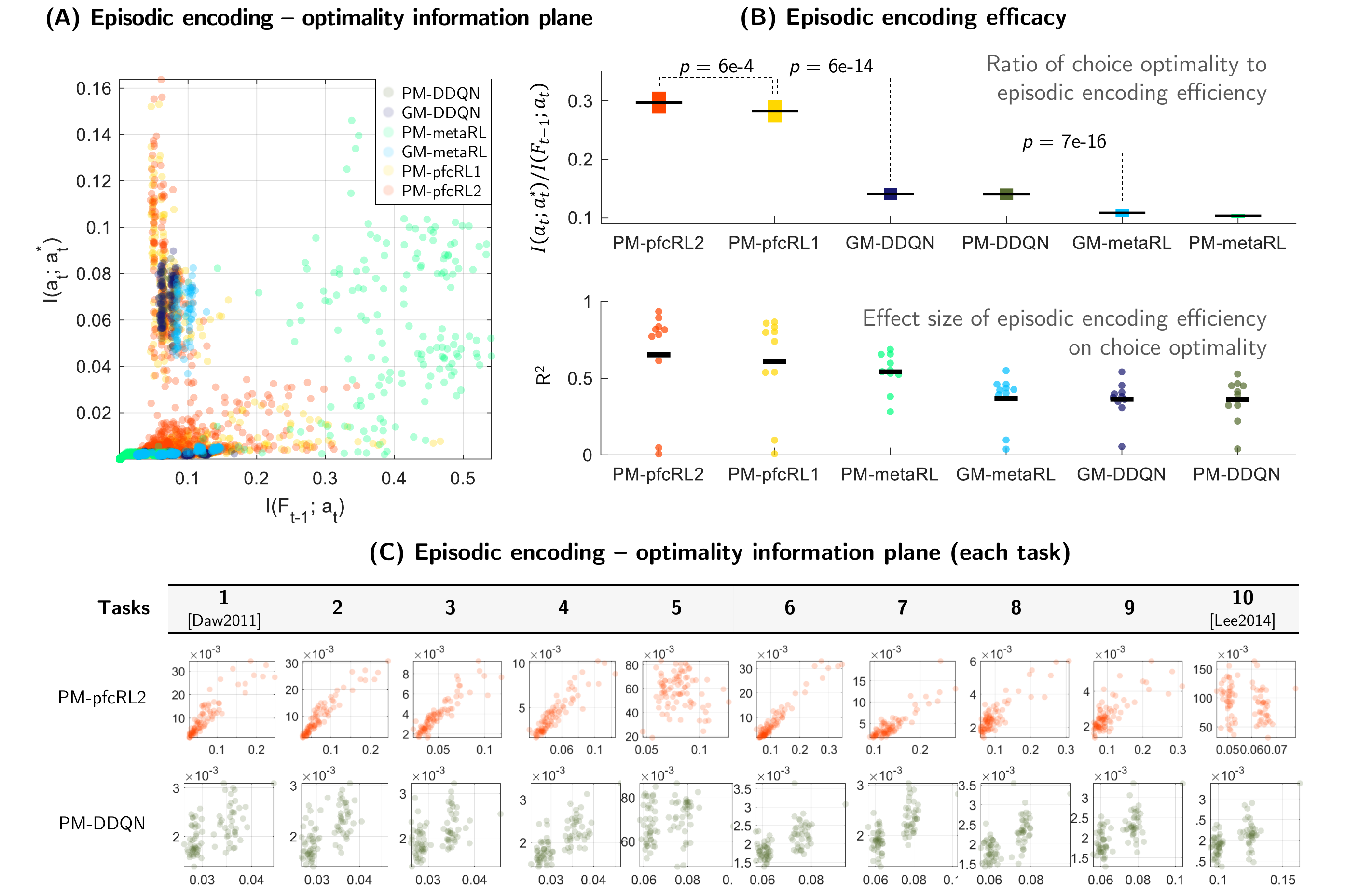}
  \caption{Episodic encoding efficacy represents the agent's ability to transfer information from past episodes to the agent's action and task performance. (A) Episodic encoding efficiency-optimality information plane. The X-axis shows the mutual information of the episodic events and the agent's action (``episodic encoding efficiency''). The Y-axis is the mutual information of the agent's action and the optimal action (``choice optimality''). (B) Episodic encoding efficacy. It is defined as the ratio of episodic encoding efficiency and choice optimality. (C) Episodic encoding efficacy of the most generalizable (PM-pfcRL2) and least generalizable agent (PM-DDQN) across 10 tasks. }
  \label{fig:efficacy}
    \vspace{-0.5em}
\end{figure}

\textbf{Episodic encoding efficacy test.} We computed these two measures while each RL model performed 10 Markov decision tasks (Figure 6-(A)). We then used these measures to compute the ratio \(I(F_{t-1};a_t)/I(a_t;a_t^*)\) and its goodness-of-fit statistics as a proxy for episodic encoding efficacy (Figure 6-(B)). We found that the \textit{prefrontal RL} (both PM-pfcRL1 and PM-pfcRL2) exhibited the highest level of episodic encoding efficacy. Notably, the most generalizable model, PM-pfcRL2, showed a significant correlation between episodic encoding efficiency and choice optimality in 8 of our 10 tasks (Figure 6-(C)). Also note that the empirical generalizability (Figure 5) mostly matched the episodic encoding efficacy (\(R^2\) of Figure 6-(B)). For more details, refer to \textit{Supplementary Figure.}

These results have three important implications. First, the episodic encoding efficiency helps us better understand the nature of generalizability. Second, the episodic encoding efficacy can be a good candidate for quantifying the agent’s generalizability. This measure can be directly used to design highly generalizable RL algorithms.

\section{Conclusion}
\label{sec:conclusion}

This study explored the possibility that algorithms can learn generalizable policies from humans. In doing so, we ran a large-scale experiment by fitting to 82 human subjects' data (Figure 1-(A)) different types of RL models, including \textit{deep RL}, \textit{meta RL}, and \textit{prefrontal RL}. To empirically test generalizability, we ran two formal tests as prerequisites: a reliability test and an empirical generalizability test. In the reliability test to compare the latent behavior profile of the humans with that of the RL models, we showed that the computational model, adaptively combining model-based and model-free control, called the prefrontal RL, reliably learns the latent policy of human subjects, whereas all the other models failed to pass this test. In the empirically generalizability test, we showed that the prefrontal RL indeed has the ability to generalize what it learned from the original task. In the subsequent information-theoretic analysis, we found that the prefrontal RL showed that the highly generalizable RL model has better episodic encoding efficacy.

To avoid any misinterpretation, we should note that our results do not imply that deep RL or other variants cannot generalize their policy. This is mainly because all RL models were intentionally trained to learn the policy of humans as opposed to performing the task by themselves. In this regard, this suggests an interesting implication that the way in which deep RL algorithms solve tasks is distinctly different from that of humans; so, deep RL can provide new insight into human problem solving. Conversely, our framework allows for RL algorithms to glean valuable insights from human problem-solving processes.

Our work is the first attempt to formally test the possibility that computational models can reliably learn the latent policy of humans. Most importantly, our results increase optimism regarding the design of RL algorithms with human-like intelligence. Future work should concern improving efficiency in learning the latent policy of the human, exploiting the idea of episodic encoding efficacy to design highly generalizable RL algorithms.

\section*{Broader Impact}

We used a dataset of human subjects~\cite{lee2014neural,Heo2018,kim2019task} to fit our models. All the data were anonymized. There are no ethical issues since our study does not involve any actual human experiments. Only modelling and simulations were conducted using the data. This work has the following potential impacts on society. 1) Machine learning algorithm design. Our model learns its policy by mimicking the way the human solves problems. This approach can thus stimulate the development of highly generalizable reinforcement learning algorithms. The training process is based on human behavioral datasets collected independently, so it does not require any direct interaction with human subjects. Therefore, there is no psychological or physical harm inflicted by our model. 2) Assessing the efficiency of learning. This study introduced two metrics to quantify the learning efficiency: the recoverability of the human behavior profile and the information encoding efficacy. While these measures can assist people in assessing the efficiency of human reinforcement learning, one should not exploit this model to either mislead people or discriminate unfairly against anyone. 3) Human-computer interaction. Our approach helps us design algorithms that can learn like humans, making it possible to design human-friendly human-computer interaction systems. That said, one should not misread this and think that it can guide or replace human-human interactions.



\bibliography{ms}
\bibliographystyle{plain}




\end{document}


\title{{\Large Supplementary Materials for}\\
\vspace{1.0em}On the Reliability and Generalizability of Brain-inspired Reinforcement Learning Algorithms}
\maketitle
\tableofcontents
\listoffigures

\let\thefootnote\relax\footnotetext{The simulation codes are available at https://github.com/brain-machine-intelligence/simul-mdps}

\newpage

%

\section{Building RL models that learn the latent policy of humans}
\label{subsec:building rl}
\vspace{-1.2em}
\textbf{\hspace{0.2in}\large(related to Section 2 of the main text)}
\vspace{1.2em}

We used the following three types of reinforcement learning (RL) models for the simulations: DDQN~\cite{van2016}, meta RL~\cite{wang2016learning, wang2018meta}, and prefrontal RL~\cite{lee2014neural, kim2019task}. The first two models were trained using two different training methods, goal matching (GM) and policy matching (PM). The prefrontal RL models were trained using the PM method only. In total, six RL models were trained to conduct the large-scale simulations for both the recoverability test and the generalizability test (also see Figure 2 in the main text). For the initial training, we used the two-stage Markov-decision task (MDT)~\cite{lee2014neural} as an environment.

\subsection*{The environment}

The two-stage MDT was used as an environment~\cite{lee2014neural} for each RL model to train. It was originally designed for human subject experiments with regard to choice behavior. The structure of the task was designed to optimally favor behavioral control by either the model-based or model-free learning strategy. On each trial, it requires sequential binary choices through a two-layered Markov-decision problem (MDP) in order to obtain tokens that are redeemable for a monetary reward (Figure S1). Each trial in the task requires two sequential actions. 

The task has mainly two types of trials – specific and flexible goal. In specific goal trials, an agent can obtain a monetary reward when the agent chooses only one color of tokens. The color of the tokens allowing the monetary reward is changed on a trial-by-trial basis. For example, if a blue color was given with the specific goal condition, an agent should collect the blue-colored token to obtain a monetary reward. In contrast, in flexible goal trials, an agent can obtain a monetary reward when the agent collects any color of tokens. In the specific goal condition, a model-based strategy is favored since the reward prediction error (PE) will be high on average due to constant changes in goal state-values, whereas a model-free strategy is more encouraged in flexible goal trials.  

In addition to this manipulation of goal conditions, the task also manipulated state-action-state transition probabilities within the MDP. Therefore, uncertainty in state-action-state transitions is high (0.5 vs. 0.5) on some occasions and becomes low (0.9 vs. 0.1) on other occasions. This manipulation of state-action-state transition probabilities across the trials intends to elicit either high or low state- PEs on average that should favor either a model-free or a model-based strategy, respectively.

\begin{figure}[t!]
  \includegraphics[width=\textwidth]{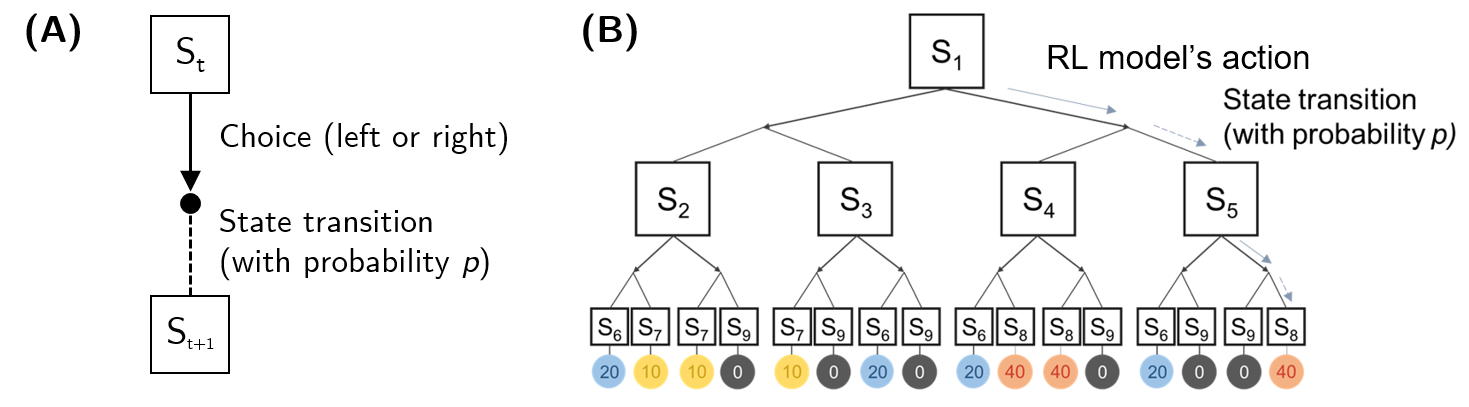}
  \raggedright 
  \caption{Two-stage Markov-decision task (MDT)~\cite{lee2014neural}.}{(A)  Sequential two-choice Markov decision task ([4], redrawn). An agent moves from one state (St) to the other state (St+1) with a certain state-transition probability p following a binary choice, either left or right. (B) Illustration of the task.}
  \label{fig:mdt}
\end{figure}

\subsection*{A Markov-decision process subject to the training method }

With the aforementioned two-stage MDT, we formalize the choice behavior of agents as an MDP, which is a 5-tuple (\(S, A, T, R, \gamma\)), where \(S\) is the state space, \(A\) is the action space, \(T\) is the transition function, \(R\) is the reward function, and \(\gamma\in[0, 1)\) is the discount factor. Since the GM and the PM have different objectives (and associated goal conditions), the specific definition of the MDP varies depending on the training method.  

\textbf{Goal matching (GM)}. In this training method, the RL model interacts with the two-stage MDT to simply maximize the expected amount of reward. We applied this GM method to two RL models, DDQN~\cite{van2016} and meta RL~\cite{wang2016learning, wang2018meta}, which we called GM-DDQN and GM-metaRL hereinafter, respectively. The MDP is specifically as follows:

\begin{itemize}[leftmargin=0.3in]
\item State: The state \(S\) is defined as the combination of the current state in the task and the current goal condition. The current state in the task is defined as S1, S2 … S9, as shown in Figure S1. The goal condition includes flexible and specific goal types.

\item Action: The action space \(a_t \in A\) of state \(s_t \in S\) is defined as two actions, moving either to \textit{Left} or to \textit{Right} at state \(s_t\). This action space allows the agent to move to the next state to get a token redeemable for the monetary reward subject to the goal condition.

\item Transition: \(T:S \times A \times S \rightarrow [0,1]\) is a transition function driven by the environment \(E\). Given the state \(s_t\) and an action \(a_t\), the transition to the next state \(s_{t+1}\) is determined by the state-transition probability in each trial, either (0.9 vs. 0.1) for the low-uncertainty condition or (0.5 vs. 0.5) for the high-uncertainty condition.

\item Reward: \(R:S \times A \times R \rightarrow R\) is the reward function, where \(R\) is a discrete set of possible rewards in a range of \{0, 10, 20, 40\}. The reward that the RL agent actually receives varies depending on the goal condition given on a trial-by-trial basis. As described, an agent receives a (monetary) reward when the agent satisfies the goal condition in each trial. For example, on specific goal trials, the agent receives the reward if the agent successfully collected the token whose color is the same as the color that the specific goal condition indicated. On flexible goal trials, the agent receives the reward regardless of the color of tokens.
\end{itemize}

\textbf{Policy matching (PM).} In this training method, an RL agent was trained in such a way that it mimics the way the human performs reward maximization while performing a two-stage MDT. Thus, this method enables the achievement of both goal matching and behavior matching through combining the GM and behavior cloning (also see Figure 1 in the main text). In each training epoch, the RL model completes one episode consisting of 200--400 games of the two-stage MDT to maximize the reward (GM), and then the discrepancy between the model’s behavior and human subject's behavior is translated into the loss function (behavior cloning). Here, the actual training of the RL model is guided by the loss function constructed by the difference between the agent's and the human subject's behavior, not  by the reward acquired while performing each game of the two-stage MDT. Therefore, we present the new reward of the MDP in PM, a fundamental difference between GM’s and PM’s MDP.

\begin{itemize}[leftmargin=0.3in]
\item Reward: \(R:S \times A \times R \rightarrow R\) is the reward function, where \(R\) is a discrete set of possible rewards in a range of \{\(k-n, k, k+n\)\}, \(k>0, n \ge 0\). The terminal reward \(R_\Omega\) is defined as
\begin{equation}
   R_\Omega=
   \begin{cases}
   k-n, & \text{if}~a_{ag}^1 \neq a_H^1~\text{and}~a_{ag}^2 \neq a_H^2 \\
   k+n, & \text{if}~a_{ag}^1=a_H^1~\text{and}~a_{ag}^2=a_H^2 \\
   k,    & \text{otherwise} 
   \end{cases},
\end{equation}

where \(k, n\) is a pre-defined constant value, \(a_{ag}^i\) is an action that the agent performed at stage \(i\) in the two-stage MDT, and \(a_H^i\) is an action that a human subject originally performed at the same stage \(i\). 
\end{itemize}

The practical meaning of \(R_\Omega\) is that an RL model is able to receive the maximum reward (e.g., \(k+n\)) when the RL model can duplicate the choice behavior of the human subject in one game. If the RL model completely fails to the duplication of the human subject's behavior in one game, the RL model receives the minimum reward (e.g., \(k-n\)). The RL model receives a neutral reward (e.g., \(k\)) otherwise. Here, the amount of n determines the impact of the terminal reward \(R_\Omega\) on the reinforcement of the RL model's policy.
 
We note that to determine the terminal condition, we compared only two actions of the RL model and the human subject's actions since the two-stage MDT environment requires sequential two choices. For the entire training using PM, we ran 1,000 epochs and 8,000 epochs for the training of PM-DDQN and PM-metaRL, respectively.

\subsection*{Training}

Given the two training methods, we trained the following six RL models: DDQN and meta RL using GM (GM-DDQN, GM-metaRL), DDQN and meta RL using PM (PM-DDQN, PM-metaRL), and prefrontal RL using PM (PM-pfcRL1, PM-pfcRL2). 

DDQN is a deep reinforcement learning algorithm designed to effectively maintain a thematic consistency between value and policy.

\begin{equation}
Y_t^{DDQN} \equiv R_{t+1} + \gamma Q(s_{t+1}, \text{argmax}~Q(s_{t+1},~a;\theta_k), \theta_k^-).
\end{equation}

We implemented the DDQN using the following network architecture: one input layer; two hidden layers, which were composed of 64 nodes each; and one output layer. All layers are fully connected. A set of hyper-parameters is as follows: the discount factor \(\gamma\) is 0.99, the learning rate \(\alpha\) is 0.001, the target update frequency to the primary network \(\tau\) is 0.001, and the batch size is 32. 

For meta RL, we used the A3C-LSTM model implemented by A. Juliani ( \url{https://github.com/awjuliani/Meta-RL}). The size of the LSTM is 256. The state, previous reward, and previous actions were encoded in the form of one-hot vectors. The LSTM receives an input as the concatenated form of these vectors. The policy and value networks are fully connected feedforward networks without hidden layers. These two networks receive LSTM outputs. 

For prefrontal RL, we used the computational model adaptively combining model-based and model-free control, inspired from the findings on the neural activity of the lateral prefrontal cortex and ventral striatum~\cite{lee2014neural, kim2019task}. Two versions were selected: a baseline model for PM-pfcRL1~\cite{lee2014neural} and an adaptive model for PM-pfcRL2~\cite{kim2019rldm}. Both models successfully implement arbitration control based on the Bayesian reliability estimation over model-based RL (FORWARD learning~\cite{glascher2010states}) and model-free RL (SARSA learning~\cite{sutton1998reinforcement}). PM-pfcRL2~\cite{kim2019rldm} is slightly different from the original one (PM-pfcRL1). PM-pfcRL2 computes Bayesian reliability based on the clusters of PE distribution estimated by the Dirichlet process Gaussian mixture model, while PM-pfcRL1 simply divides PEs with the pre-defined threshold. By considering the PE distribution to be a Gaussian mixture model, PM-pfcRL2 is able to incorporate dynamic changes of task variables in the environment.

In addition, we used a random agent as an agent for the control group. The random agent chose the action randomly while performing the two-stage MDP. Figure S2 shows the training results of all the RL models. 

In the GM training, we used a fixed number of episodes for all RL models, GM-DDQN and GM-metaRL to train. In the PM training, we used the early-stopping condition for PM-DDQN, and maximum number of episodes for both PM-DDQN and PM-metaRL. For the computational efficiency, we stopped the training of PM-DDQN earlier when its likelihood summation in a single episode was greater than the likelihood summation of PM-pfcRL. Here, the likelihood stands for the probability showing the extent to which the choice of the agent at state \(s_t\) is similar to the actual choice of the human subject at the same state \(s_t\). We followed equation~\ref{eq:likelihood} to compute the likelihood,

\begin{equation}
\text{Likelihood}~L = p(a|s_i),
\label{eq:likelihood}
\end{equation}

where \(a\) is the human subject's choice at the \(i^{th}\)  trial in state \(s\).

Supplementary Figure S2 presents the training results of all the RL models. In Figure S2A, a family of prefrontal RL models (PM-pfcRL1, PM-pfcRL2) shows the best performance in the aspect of rapid learning. It shows the consistent performance from the beginning stage of training to the end at a relatively higher level. The deep RL model, PM-DDQN, shows the worst performance. Rapidly decreased at the beginning, it shows a consistently low-level performance. Perhaps this gives us an intuition that a purely model-free RL agent may not be able to learn the optimal policy via the PM method. In the case of meta RL, PM-metaRL presents an interesting pattern – it learns slowly but can achieve the best performance in the end out of four RL models. Figure S2B shows the training results of the RL models in which GM was used: GM-DDQN and GM-metaRL.

\begin{figure}[t!]
  \raggedright 
  \includegraphics[width=\textwidth]{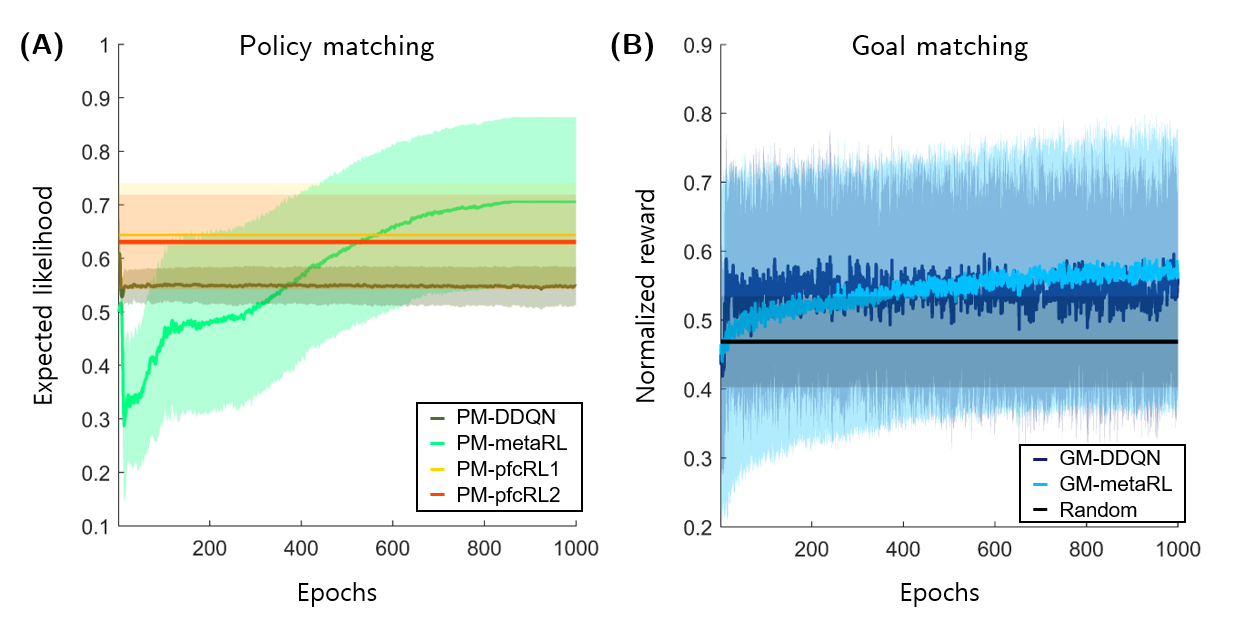}
  \caption{Training results of all RL models.}{(A) Training result of RL models trained using PM. The X-axis represents the number of epochs, and the Y-axis represents the expected likelihood, meaning the extent to which the RL model accurately predicts the action at the state compared to that of the human subject at the same state. (B) Training result of RL models trained using GM. The X-axis represents the number of epochs, and the Y-axis represents the normalized reward that each RL model trained using the GM method received.}
  \label{fig:training}
\end{figure}

\newpage

\section{Latent behavior profile recovery test}
\vspace{-1.0em}
\textbf{\hspace{0.2in}\large(related to Section 3 of the main text)}
\vspace{1.0em}

\subsection*{General linear model (GLM) for the latent behavior profile recovery test}

We used the GLM to conduct the parameter recovery analysis on choice behavior. Here, the choice optimality, which is the behavioral measurement of model-based control~\cite{kim2019task}, was chosen for a dependent variable, and uncertainty, complexity, previous reward, previous action, and max goal value were chosen for independent variables. Thus, the model was,

\begin{equation}
y=\beta_1 x_1+\beta_2 x_2+\beta_3 x_3+\beta_4 x_4,
\end{equation}

where \(y\) is a choice optimality, and \(x=\{x_1:\text{Uncertainty},~x_2:\text{Goal},~x_3:\text{Action (Prev.)},~x_4:\text{State (Prev.)}\}\).

\section{Testing empirical generalizability of the RL models}
\vspace{-1.0em}
\textbf{\hspace{0.2in}\large(related to Section 4 of the main text)}
\vspace{1.0em}

In order to test the capacity of the three types of RL models (see also Figure 2 in the main text) to generalize what they learned from a single task, we situated those RL models already trained in the two-stage MDT in the new volatile environments that the RL models never experienced. We created ten different MDTs for the environments, where the following two task parameters can be manipulated: task structure (ladder and tree) and task uncertainty (fixed, drift, switch, and drift + switch).

\subsection*{Testing the generalizability of the RL models trained using PM} 

We used the following four RL models for this scenario: PM-DDQN, PM-metaRL, PM-pfcRL1, and PM-pftRL2. First, we trained these four RL models in the two-stage MDT using the PM training method, which resulted in a total of 328 RL models (82 human subjects' data \(\times\) 4 RL models). We saved the entirety of the parameters and weights of each RL model once all the RL models had been trained. To measure the empirical generalizability of the RL models, we situated the restored RL models in the ten tasks. The order of tasks that the RL agent experiences was pseudo-randomly determined. Once one RL model was loaded, the RL model performed the task assigned under the condition that both the parameters and weights of the model were frozen. The sequence on the manipulation of task variables (e.g., state-transition uncertainty and goal condition) were fixed across the entirety of the  RL models involving the simulations.  

\subsection*{Testing the generalizability of the RL models trained using GM} 

We used the following two RL models for this scenario: GM-DDQN and GM-metaRL. Here, the two RL agents were allowed to perform all tasks pseudo-randomly, given without any constraints. Once the two RL models had been successfully trained in all ten tasks, the RL models followed the same steps as in the case of PM. After restoring all parameters and weights of each model, the RL model performed the task assigned under the condition that both the parameters and weights of the model were frozen.

\begin{figure}[t!]
  \raggedright 
  \includegraphics[width=\textwidth]{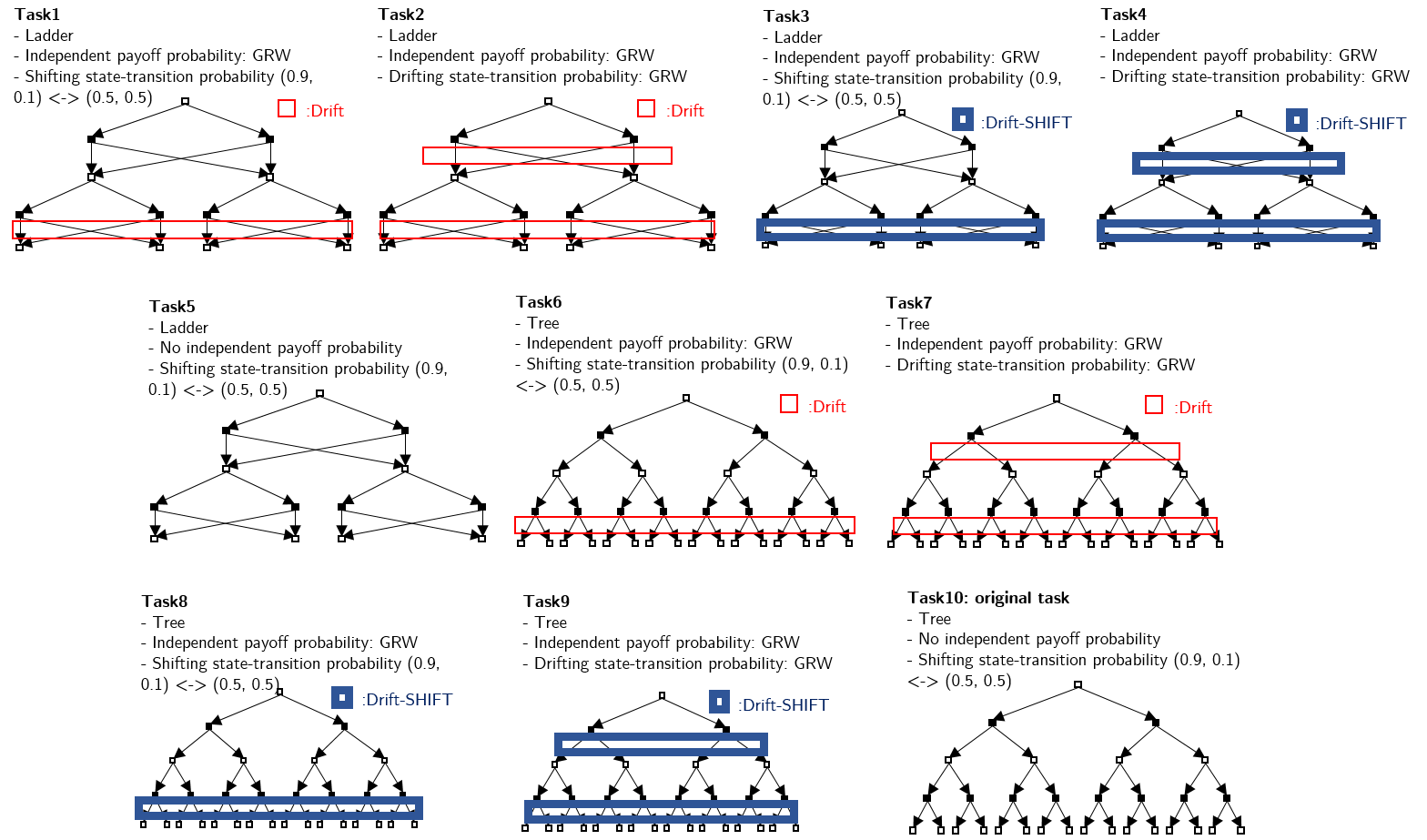}
  \caption{Ten MDTs used for the empirical generalizability test (see also Figure 4 in the main text).}
\end{figure}

\newpage

\begin{figure}[t!]
  \raggedright 
  \includegraphics[width=\textwidth]{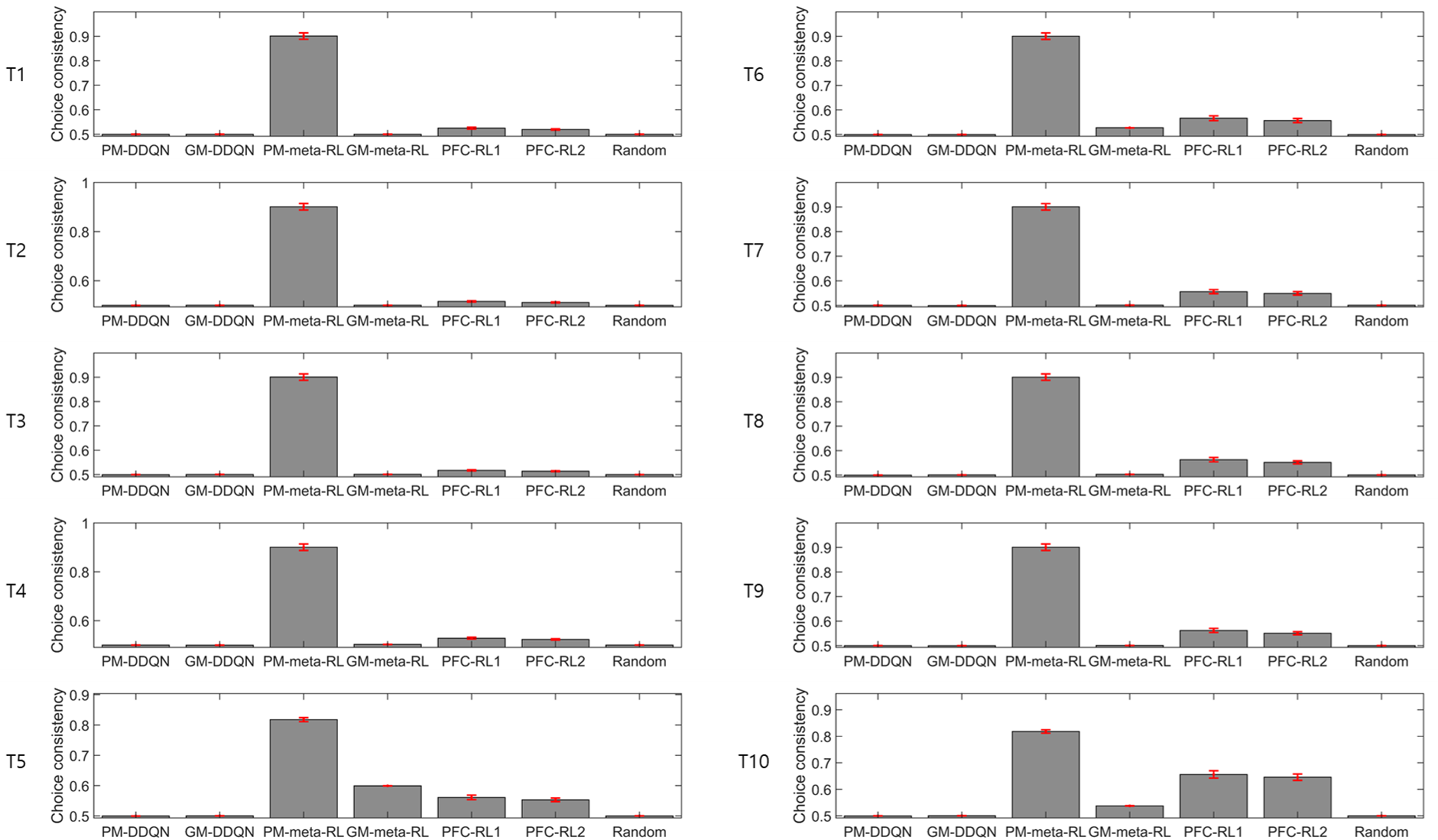}
  \caption{Behavior profile – Choice consistency.}{The X-axis represents the RL model, and the Y-axis represents choice consistency while each RL model performs the task. The label on the left side of the Y-axis represents the index of the task, e.g., T1 refers to Task 1 as presented in Figure S3.}
\end{figure}

\begin{figure}[t!]
  \raggedright 
  \includegraphics[width=\textwidth]{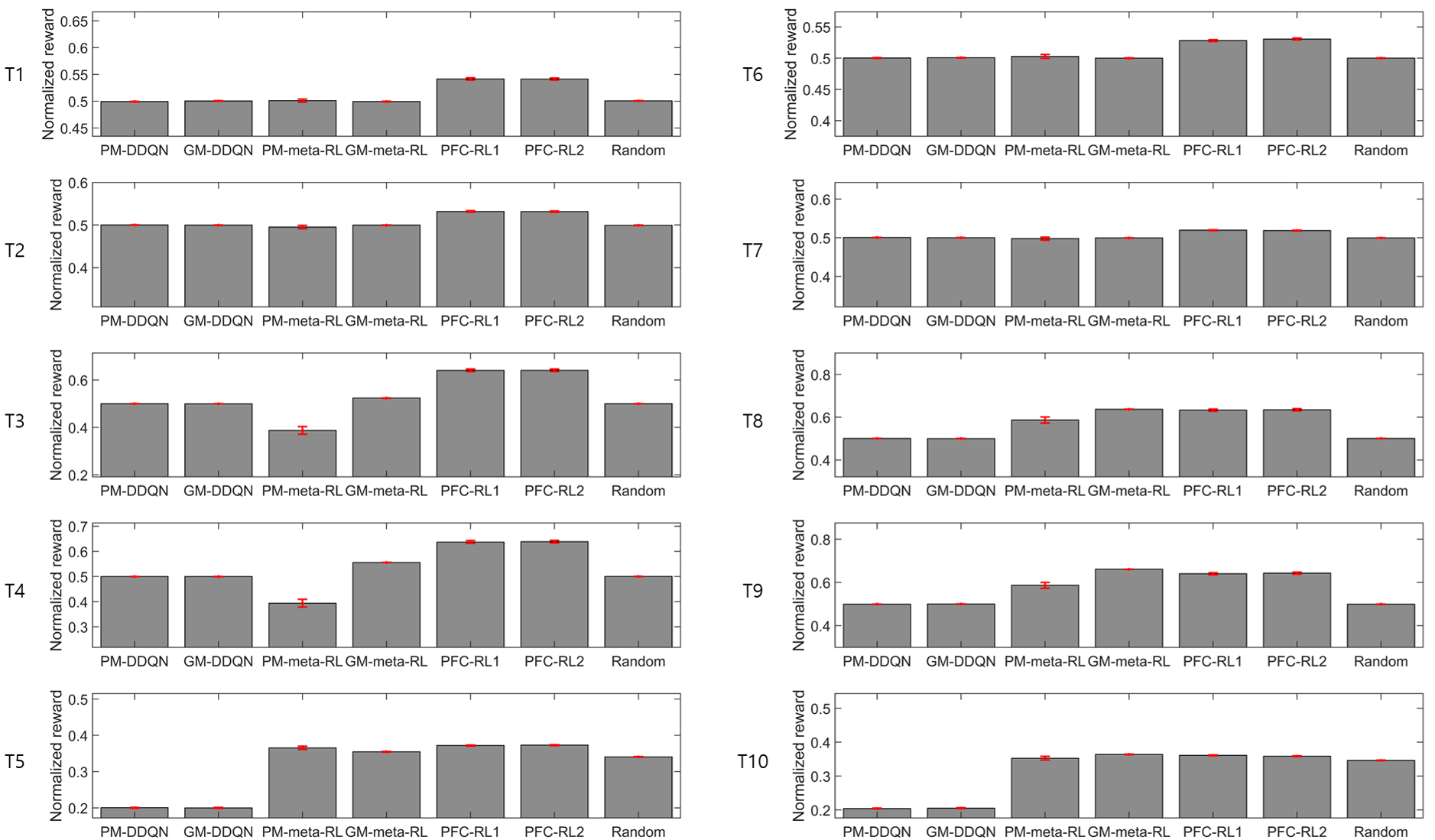}
  \caption{Behavior profile – Normalized reward.}{The normalized reward when model changed an action (i.e., \(a_{t-1} != a_t)\). The X-axis represents the RL model, and the Y-axis represents a normalized reward while each RL model performs the task. The label on the left side of the Y-axis represents the index of the task, e.g., T1 refers to Task 1 as presented in Figure S3.}
\end{figure}

\clearpage
\newpage

\section{Episodic encoding efficacy test via information theoretic analysis}
\vspace{-1.0em}
\textbf{\hspace{0.2in}\large(related to Section 4 of the main text)}
\vspace{1.0em}

The information theoretic analysis is designed to measure the mutual information between two measures collected while the human subjects participated in the task on choice behavior~\cite{Filipowicz2019}. Here we used episode \(F\), defined as,

\begin{equation} 
F_t=\{a_{t-1}^2, S_{t-1}^3, a_{t-1}^{2,*}, R_{t-1}\},
\end{equation}

where 
\(a_{t-1}^2\) is an action that was taken by an agent at stage 2 in the previous trial, \(S_{t-1}^3\) is a rewarding state at stage 3 in the previous trial,  \(a_{t-1}^{2,*}\) is an optimal action at stage 2 in the previous trial, and \(R_{t-1}\) is a reward in the previous trial.

In this analysis, we computed \(I(F_t,a_t)\), which stands for the level of compression. It measures the amount of mutual information between the current action and the history of the previous trial. In addition, we computed \(I(a_t,a_t^*)\), which measures the optimality of the choice behavior~\cite{Filipowicz2019}. 

We call \(I(F_t, a_t)\) the episodic encoding efficacy. We assume that this quantifies the amount of information required to achieve the optimal choices by the agent. It is direct measure of the agent's capacity to mimic human-like behavior in relation to the governing of the trade-off between performance and efficiency under the limited cognitive resources.

\begin{figure}[t!]
  \raggedright 
  \includegraphics[width=\textwidth]{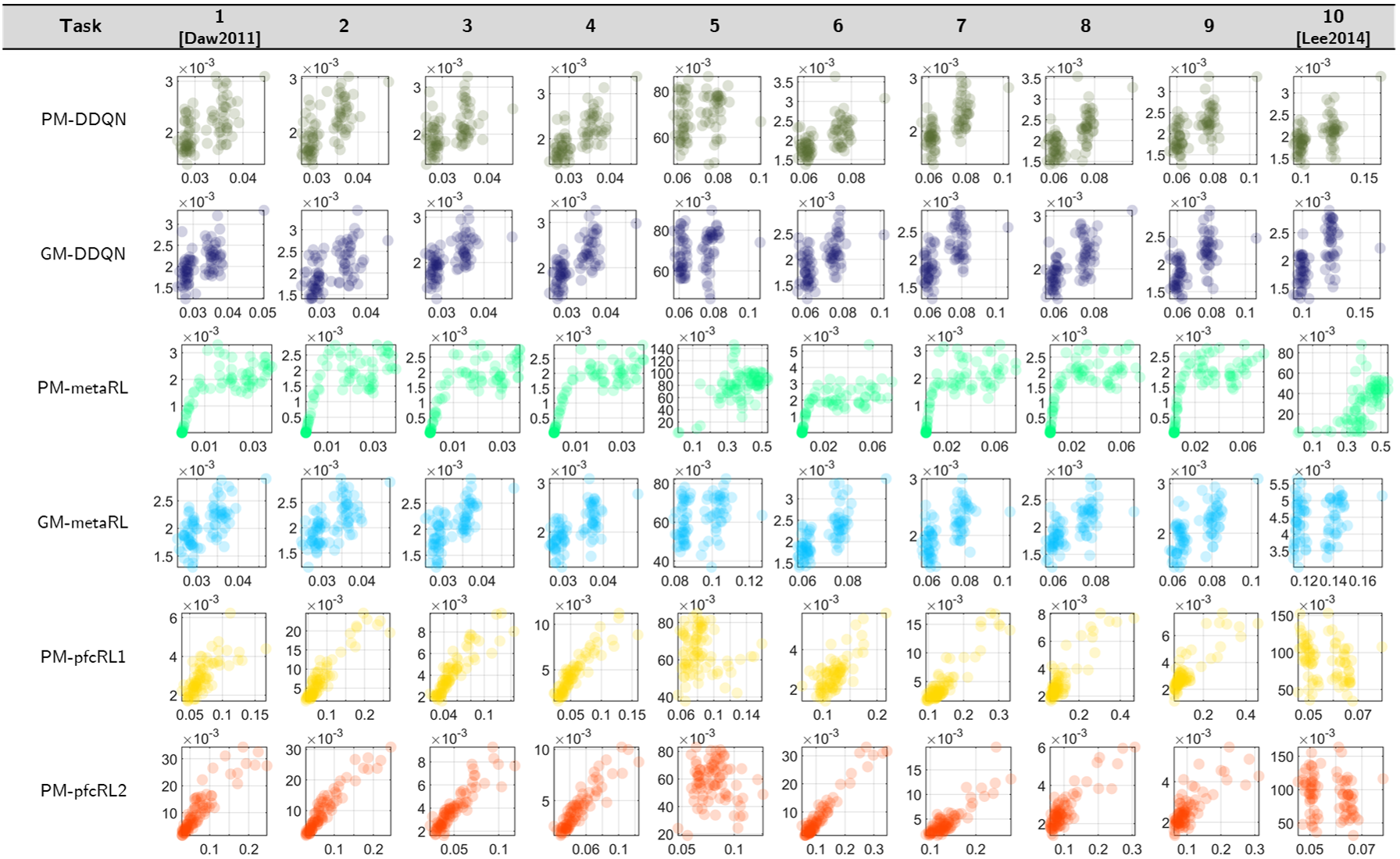}
  \caption{Episodic encoding efficacy for six RL models across ten tasks.}{Each plot shows an episodic encoding efficiency-optimality information plane. It represents the agent’s ability to transfer information from past episodes to the agent's action and task performance. The X-axis shows the mutual information of the episodic events and the agent's action (``episodic encoding efficiency''). The Y-axis is the mutual information of the agent’s action and the optimal action (``choice optimality''). In the table, the higher the position in the row, the worse capacity to generalize what the agent learned from a single task. For example, PM-DDQN is the worst, and PM-pfcRL2 is the best.}
\end{figure}



\bibliography{supplement}
\bibliographystyle{plain}


